\documentclass{article}
\usepackage{spconf,amsmath,graphicx}

\usepackage{enumitem}
\setlist{nosep, leftmargin=14pt}

\usepackage{mwe} 
\usepackage{amssymb}
\usepackage{multirow}
\usepackage{bbding}
\usepackage{cite}
\usepackage{setspace}

\title{Enhancing Representation in Medical Vision-Language Foundation Models via Multi-Scale Information Extraction Techniques}

\name{%
\begin{tabular}{c}
    {Weijian Huang$^{1,2,3}$ \qquad Cheng Li$^{1}$ \qquad Hong-Yu Zhou$^{4}$ \qquad Jiarun Liu$^{1,2,3}$} \\
    {Hao Yang$^{1,2,3}$ \qquad  Yong Liang$^{2,5}$ \qquad Guangming Shi$^{2}$ \qquad Hairong Zheng$^{1}$ \qquad Shanshan Wang$^{1,\dagger}$\thanks{$^{\dagger}$ Corresponding author. ss.wang@siat.ac.cn}}
  \end{tabular}
}
\address{$^{1}$Paul C. Lauterbur Research Center for Biomedical Imaging, \\
Shenzhen Institute of Advanced Technology, Chinese Academy of Sciences, Shenzhen, China\\
$^{2}$Peng Cheng Laboratory, Shenzhen, China\\
$^{3}$University of Chinese Academy of Sciences, Beijing, China\\
$^{4}$Department of Computer Science, The University of Hong Kong, Pokfulam, China\\
$^{5}$Pazhou Laboratory (Huangpu), Guangzhou, China
}

\begin{document}
%
\maketitle
\begin{abstract}
The development of medical vision-language foundation models has attracted significant attention in the field of medicine and healthcare due to their promising prospect in various clinical applications. While previous studies have commonly focused on feature learning at a single learning scale, investigation on integrating multi-scale information is lacking, which may hinder the potential for mutual reinforcement among these features. This paper aims to bridge this gap by proposing a method that effectively exploits multi-scale information to enhance the performance of medical foundation models. The proposed method simultaneously exploits features at the local, instance, modality and global aspects, facilitating comprehensive representation learning within the models. We evaluate the effectiveness of the proposed method on six open-source datasets across different clinical tasks, demonstrating its ability to enhance the performance of medical foundation models.

\end{abstract}
\section{INTRODUCTION AND RELATED WORK}
\label{sec:intro}
Recent advancements in machine learning have greatly improved automated diagnostic systems (ADS), achieving expert-level performance\cite{chang2023mining, wang2021annotation, zhou2019d, zhou2023transformer,zhou2023unified}. In the domain of medical imaging, medical foundation models have emerged as a highly promising method \cite{krishnan2022self}. These models can adapt to diverse clinical applications while minimizing the reliance on extensive downstream annotations \cite{tang2022self}.

To enhance the expressive capabilities of the foundation models, researchers are progressively incorporating multimodal data, such as clinical reports derived from routine examinations, alongside the radiography \cite{REFERS, MRM, BIOVIL, GLORIA, M-FLAG , MEDKLIP, CONVIRT, MED-UNIC, MaCo}. This vision-language models aim to capture a wealth of expert knowledge and improve the models’ understanding of the medical context, leading to more comprehensive and accurate representations. To achieve this, common strategies involve leveraging contrastive learning to align paired multimodal data and disentangle the feature distributions of unpaired data, thereby improving the representation capability of the image encoder \cite{CLIP, MEDKLIP, CONVIRT}. 

However, since medical data has many properties, including fine-grained semantic understanding, radiography-specific clinical reports, and the inherent properties of single-modality representation, several crucial factors necessitate further consideration. Firstly, on a local scale, it is commonly observed that each semantic within a medical report corresponds to distinct and independent local areas within the image. To ensure precise and comprehensive representation, it is crucial to effectively establish the spatial correspondence between each semantic and specific regions of the image \cite{GLORIA}. Secondly, at the instance scale, the report and corresponding image typically exhibit a unique correspondence compared to other images. Neglecting the individual characteristics of each image-language pair may lead to suboptimal performance and incomplete representation learning \cite{ALBEF}. Thirdly, at the modality scale, both images and reports inherently contains distinct medical semantics that should be independently extracted and exploited. Disregarding these unique features of each modality may hinder the comprehensive expression of the foundation model in downstream tasks, limiting its effectiveness and clinical applications.

As a result, relying solely on single scale learning may limit the performance of the learned foundation models \cite{GLORIA, MGCA}. Some methods have been proposed to address this issue by extracting features at different scales\cite{GLORIA, MGCA}. For example, Gloria \cite{GLORIA} utilizes word-scale responses to capture local matching representations. However, relying solely on individual words may not fully represent a specific region adequately. At the instance scale, ALBEF \cite{ALBEF} considers image-text matching pairs to enhance model performance. However, it relies on a large transformer for modal fusion, which raises concerns regarding its applicability in clinical practice. At the modality scale, MAE \cite{MAE} and Bert \cite{BERT} propose to use self-supervised masked training to obtain unique representations of images/text. However, they do not consider the multimodal context in medical environments. Improved approaches such as MRM \cite{MRM} combine the mechanisms of masking but sacrifice important capabilities like zero-shot learning due to the fusion of modalities, thereby limiting their application in certain scenarios. More importantly, to the best of our knowledge, there is currently no method that simultaneously takes into account the aforementioned multi-scale information, potentially overlooking the mutual enhancement relationships among these different scale features.

In light of these observations, this paper develops a multi-scale feature learning framework for radiography-reports medical foundation models, taking into consideration four different scale of information: global-, local-, instance- and modality-scale feature learning. The proposed method has been validated on six open-source datasets, covering four clinical tasks, including classification, segmentation, zero-shot classification, and phase grounding. The experimental results comprehensively demonstrate the effectiveness of the proposed method in enhancing the performance of medical foundation models.

\section{Method}
The overall methodology is illustrated in Fig. 1. In the following, we will describe the proposed multi-scale feature learning framework in detail, concentrating on the extraction and utilization of features across four distinct scales within our framework.

\begin{figure}[bht]
  \centering
  \centerline{\includegraphics[width=8.5cm]{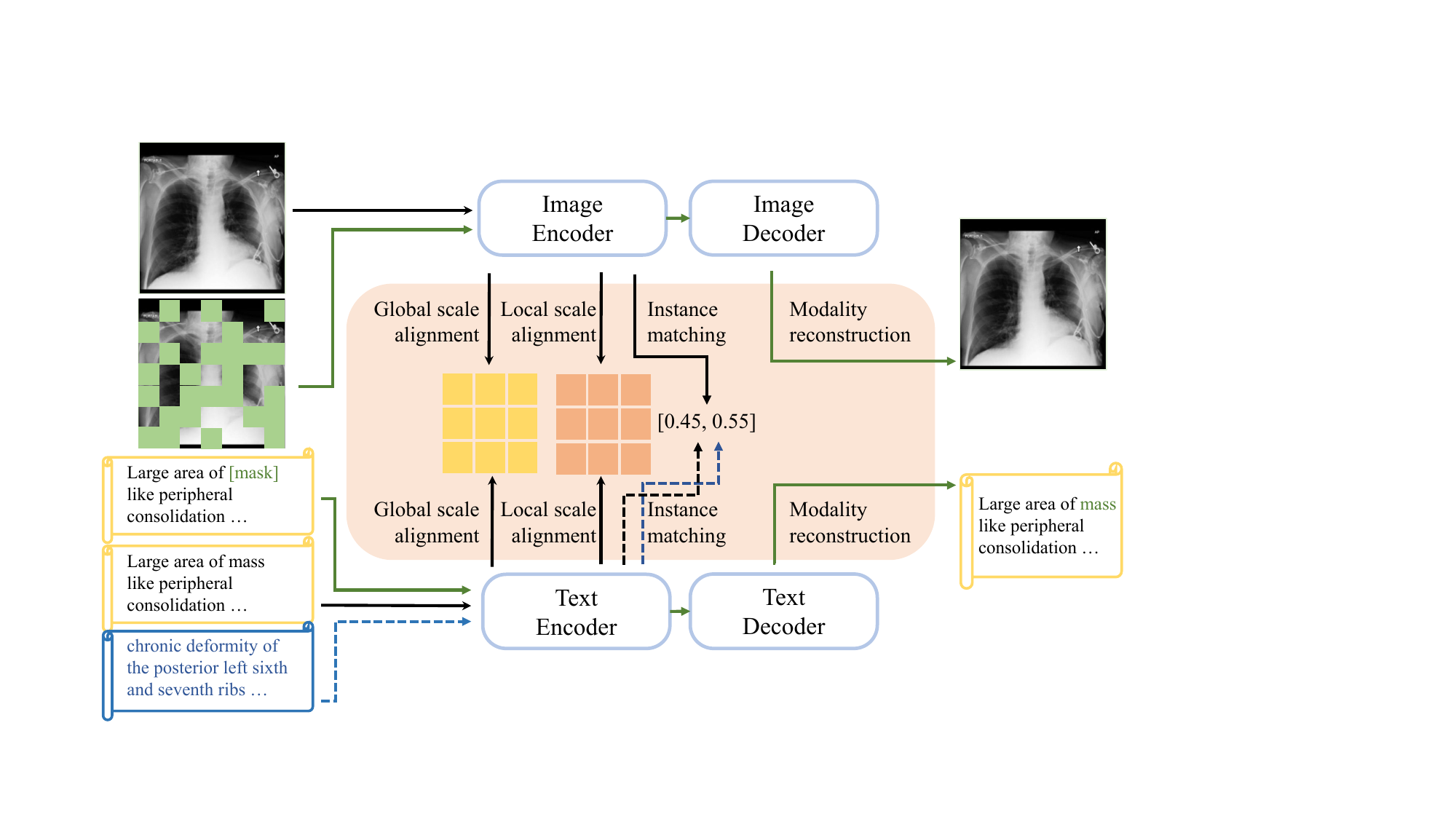}}
  \caption{The proposed foundation model. Four scales of features are simultaneously exploited to enhance the representation learning of the model.}
  \medskip
\end{figure}

\subsection{Global scale alignment}
Alignment based on complete image and text is usually treated as global scale alignment. It is an important technology in multimodal research, aiming at aligning the distributions of paired images and text representations and separating unpaired samples. Inspired by existing approaches \cite{CONVIRT, GLORIA, BIOVIL}, we employ a CLIP-base framework \cite{CLIP} to obtain the global scale feature through contrastive learning. Let $\langle v,t \rangle$ represents the cosine similarity between the visual representation $v\in \mathbb{R}^{B\times C}$ and the report representation $t\in \mathbb{R}^{B\times C}$, the global scale alignment loss $\mathcal{L}_{g}$ can be written as:
\begin{align}
\centering
\mathcal{L}^{vt}_{g} = -\sum_{i}^{B}log&(\frac{exp(\langle v_i,t_i \rangle/\tau_{1})}{\sum_{k}^{B}exp(\langle v_i,t_k\rangle/\tau_{1})}) \nonumber\\ 
\mathcal{L}^{tv}_{g} = -\sum_{i}^{B}log&(\frac{exp(\langle t_i, v_i \rangle/\tau_{1})}{\sum_{k}^{B}exp(\langle t_i, v_k\rangle/\tau_{1})}) \nonumber
\end{align}

\begin{equation}
\centering
\mathcal{L}_{g} = \mathcal{L}^{tv}_{g} + \mathcal{L}^{tv}_{g}
\end{equation}

where $B$ is the batch size and $\tau_1$ is a learnable parameter initialized to 0.07 in our experiments. 

\subsection{Local scale alignment}
Each medical sentence in a report can typically correspond to an independent local region in the image. It is crucial to effectively capture the local scale correspondence between sentence information and image regions to ensure accurate and comprehensive representations. GLoRIA \cite{GLORIA} introduces a local alignment framework that contrasts image sub-regions with words in the paired report, leading to associations between words and their corresponding image regions. However, relying solely on individual words extracted from medical text reports may be insufficient to capture the original semantic information, potentially resulting in erroneous associations. In contrast, a sentence is often a better choice. Therefore, we propose an improved approach that treats sentences as the fundamental units instead of words. Let $N_t, N_v$ denote the number of divided report's sentences and the number of divided image patches, respectively. We use a localized feature matching function $Z$ \cite{GLORIA} to aggregate the similarities between all sentence representation $t^{l}\in \mathbb{R}^{B\times N_t \times C}$ and their image weighted representation $v^{l}\in \mathbb{R}^{B\times N_v \times C}$, the local loss is then define as:

\begin{align}
\centering
\mathcal{L}^{vt}_{l} = -\sum_{i}^{B}log&(\frac{Z(v^{l}_{i},t^{l}_{i})/\tau_{2})}{\sum_{k}^{B}exp(Z(v^{l}_{i},t^{l}_{k})/\tau_{2})}) \nonumber\\ 
\mathcal{L}^{tv}_{l} = -\sum_{i}^{B}log&(\frac{exp(Z(v^{l}_{i},t^{l}_{i})/\tau_{2})}{\sum_{k}^{B}exp(Z(v^{l}_{k},t^{l}_{i})/\tau_{2})}) \nonumber
\end{align}

\begin{equation}
\centering
\mathcal{L}_{l} = \mathcal{L}^{vt}_{l} + \mathcal{L}^{tv}_{l}
\end{equation}

\subsection{Instance scale matching}
In clinical practice, each instance consists of a distinct pair of image and report. Considering the unique matching relationship at the instance aspects can be advantageous as it further distinguishes positive and negative sample pairs of text and image through modality fusion. This leads to a more robust feature distribution, enhancing the overall reliability of the approach. ALBEF \cite{ALBEF} employs an image-text-matching (ITM) approach to discriminate positive and negative samples after fusing them with a multi-modal transformer. In contrast to using a complex multi-modal transformer, we adopt a simpler approach of summarizing the features of the two modalities. We  followed by binary classification using two linear layers. The instance scale matching loss function $\mathcal{L}_{im}$ is defined as:

\begin{equation}
\centering
\mathcal{L}_{im}= -y\cdot log(f^{im}) + (1-y)\cdot log(1-f^{im})
\end{equation}
where $f^{im}$ denotes the feature representation obtained by concatenating the sampled $(v, t)$ pair in channel dimension and passing them through a linear classifier. The input of image and text may not necessarily be matched, while $y$ denotes the corresponding matching ground truth. Here, difficult samples with higher similarity have a greater probability of being sampled, as demonstrated by ALBEF \cite{ALBEF}.

\subsection{Modality reconstruction}
Each individual modality inherently carries specific information. For instance, a medical radiography can reveal the precise location of a lesion, while a textual clinical report can reflect a physician's positive or negative judgment of a disease. Thus, extracting crucial information from each modality before combining them in a multi-modal framework is beneficial. In this study, we employ well-established self-supervised training paradigms, namely Masked Language Modeling (MLM) and Masked Autoencoder (MAE) \cite{MAE}, to extract the modality-based features from radiography and reports, respectively.

For the image modality, let $v^{recon}\in\mathbb{R}^{B \times N \times C}$ denotes the decoder's reconstruction output with the input of masked image patches, and $g^{recon}\in\mathbb{R}^{B \times N \times C}$ denotes the corresponding embeddings for the ground truth patches. $N$ and $C$ represent the number of sampled patches and the dimension of the embeddings, respectively. We use a simple mean-square-error as the MAE loss function $\mathcal{L}_{mae}$:

\begin{equation}
\centering
\mathcal{L}_{mae}=(v_{recon}-g_{recon})^{2}
\end{equation}
For text modality, we adopt a cross entropy loss denoted as $\mathcal{L}_{mlm}$. The overall modality reconstruction loss $\mathcal{L}_{mr}$ is:
\begin{equation}
\centering
\mathcal{L}_{mr}=\mathcal{L}_{mae} + \mathcal{L}_{mlm}
\end{equation}

The final loss function of the proposed method is:
\begin{equation}
\centering
\mathcal{L}=\lambda_{1}\mathcal{L}_{g} + \lambda_{2}\mathcal{L}_{l} + \lambda_{3}\mathcal{L}_{im} + \lambda_{4}\mathcal{L}_{mr}
\end{equation}
Here, $\lambda$ is a hyperparameter to balance the loss terms.

\section{EXPERIMENTS AND RESULTS}
Extensive experiments have been conducted on six open-source datasets to fully validate the effectiveness of the proposed multi-scale foundation model. 

\begin{table}[t]
\caption{Comparison of AUC scores for classification performance on three open-source datasets (CheXpert, NIH ChestX-ray and RSNA Pneumonia) with varying ratios of annotated samples. }
\centering
\setlength{\tabcolsep}{1.0pt}
\begin{tabular}{cccccccccc}
\hline
\multirow{2}{*}{Methods} & \multicolumn{3}{c}{CheXpert} & \multicolumn{3}{c}{NIH} & \multicolumn{3}{c}{RSNA} \\ \cline{2-10} 
        & 1\%  & 10\% & 100\% & 1\%  & 10\% & 100\% & 1\%  & 10\% & 100\% \\ \hline
ConVIRT & 85.9 & 86.8 & 87.3  & \_   & \_   & \_    & 77.4 & 80.1 & 81.3  \\
GLoRIA  & 86.6 & 87.8 & 88.1  & \_   & \_   & \_    & 86.1 & 88   & 88.6  \\
BioViL  & \_   & \_   & \_    & \_   & \_   & \_    & 88.1 & 88.4 & 89.1  \\
M3AE    & 86.2 & 87.3 & 87.9  & \_   & \_   & \_    & 89   & 90.8 & 92.3  \\
REFERS  & 87.2 & 88.1 & 88.2  & 76.7 & 80.9 & 84.7  & 89.4 & 91.6 & 92.7  \\
MedKLIP & \_   & \_   & \_    & 77.2 & 78.9 & 83.2  & 87.3 & 88   & 89.3  \\
MFLAG   & \_   & \_   & \_    & 62.2 & 71.6 & 78.7  & \_   & \_   & \_    \\
Ours    & \textbf{88.2}    & \textbf{88.2}    & \textbf{88.3}     & \textbf{78.6}    & \textbf{82.3}    & \textbf{84.9}     & \textbf{91.7}    & \textbf{92.1}    & \textbf{93.0}     \\ \hline
\end{tabular}
\end{table}

In the first set of experiments, we first fine-tuned the proposed foundation model on various classification datasets and conducted a comprehensive performance evaluation. We took into account different annotation ratios and compared our method to seven recent approaches. The results are listed in Table 1. Our method achieves the highest AUC score under different situations, outperforming all the seven comparison methods.

\begin{table}[h]
\caption{Comparison of Dice scores for segmentation performance on 'SIIM-ACR Pneumothorax Segmentation' dataset with varying ratios of annotations.}
\centering
\begin{tabular}{cccc}
\hline
Methods  & 1\%  & 10\% & 100\% \\ \hline
ConVIRT  & 25   & 43.2 & 59.9  \\
GLoRIA   & 35.8 & 46.9 & 63.4  \\
MGCA     & 49.7 & 59.3 & 64.2  \\
M-FLAG   & 52.5 & 61.2 & 64.8  \\
Med-UniC & 56.7 & 62.2 & 64.4  \\
Ours     & \textbf{57.8}   & \textbf{70.3}   & \textbf{84.4}    \\ \hline
\end{tabular}
\end{table}
Next, we evaluate the performance of our method on segmentation tasks. We utilized the publicly available database, "SIIM-ACR Pneumothorax Segmentation", and compared our method with five top-performing benchmark algorithms. As shown in Table 2, the quantitative segmentation results once again validate the effectiveness of our proposed method in extracting meaningful segmentation features.

\begin{table}[t]
\caption{Comparison of AUC scores for zero-shot classification on
RSNA Pneumonia datasets.}
\centering
\begin{tabular}{ll}
\hline
method      & RSNA  \\
\hline
GLoRIA      & 71.45 \\
ConVIRT     & 80.42 \\
BioViL      & 82.80  \\
CheXzero    & 85.79 \\
MedKLIP     & 86.94 \\
Ours & \textbf{90.28} \\
\hline
\end{tabular}
\end{table}
We conducted zero-shot classification on the RSNA Pneumonia dataset. The quantitative results are presented in Table 3. Benefiting from the utilization of structured text, MedKLIP achieves a high AUC (Area Under the Curve) score of 86.94. Nevertheless, our method obtains the highest score of 90.28, demonstrating the superiority of the proposed multi-scale feature mining approach in zero-shot classification.

Zero-shot phase ground experiments were also conducted. In Table 4, we present the phase grounding performance of different methods. Our proposed method surpasses all three comparison methods, achieving notably improved scores with a mIoU of 0.298 and a CNR of 1.468. This indicates that with the incorporation of multi-scale and cross-modality features, the model has gained a more detailed and granular representation. Qualitative analysis, as depicted in Fig. 2, involves visualizing examples with varying quantities and different types of diseases. The results provide evidence that our method successfully captures the fine-grained associations between radiological images and reports.

\begin{table}[h]
\caption{Comparison of zero-shot phase grounding on MS-CXR datasets.}
\centering
\begin{tabular}{lll}
\hline
Methods     & mIoU  & CNR   \\
\hline
ConVIRT     & 0.238 & 0.818 \\
GLoRIA      & 0.246 & 0.930 \\
BioViL      & 0.266 & 1.027 \\
Ours & \textbf{0.298} & \textbf{1.468} \\
\hline
\end{tabular}
\end{table}

\begin{figure}[h]
  \centering
  \centerline{\includegraphics[width=8.5cm]{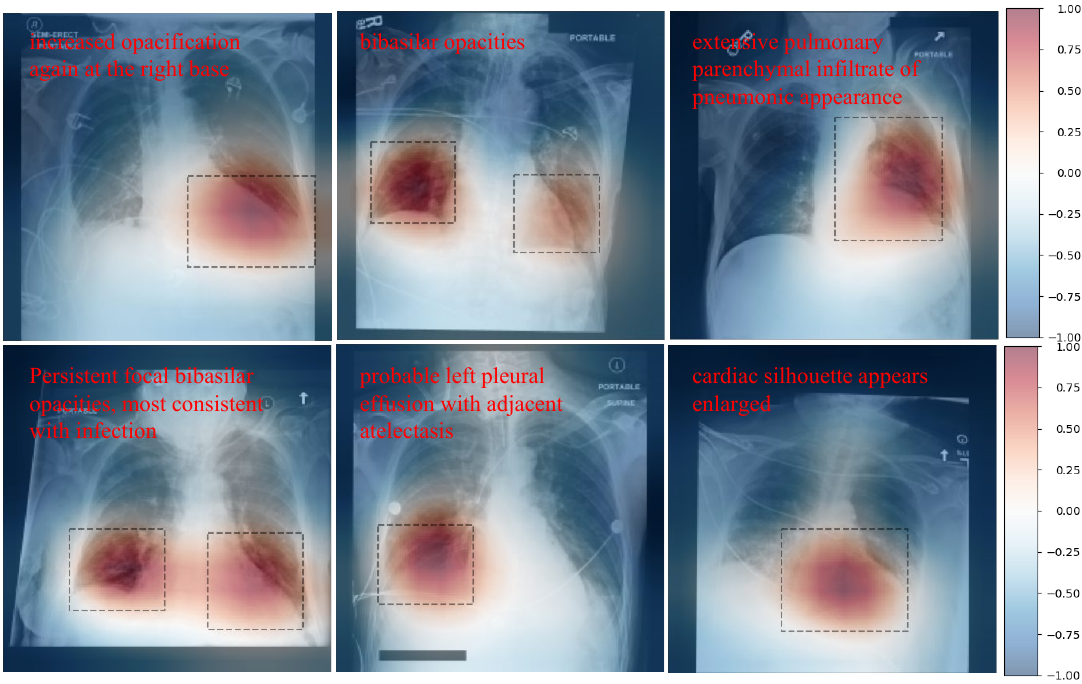}}
  \caption{Visualization of Phase Grounding. We visualize the correlation between Radiographic Images and Clinical Reports on MS-CXR Dataset. The black bounding box represents the ground truth, while deeper color of red indicates higher degree of similarity.}
  \medskip
\end{figure}

\begin{table}[]
\centering
\caption{Ablation Study. Impact of training methods with different scale on finetuning and zero-shot classfication Tasks. The comparative experiment is completed on RSNA Pneumonia dataset.}
\begin{tabular}{ccc|cc}
\hline
\multicolumn{3}{c|}{Scale}                                                        & \multicolumn{2}{c}{RSNA} \\ \hline
local                     & instance                  & modality                  & 1\% finetune & Zero-shot \\ \hline
\Checkmark &                           &                           & 91.3         & 89.5      \\
                          &\Checkmark &                           & 91.1         & 89.5      \\
                          &                           & \Checkmark & 91.2         & 85.4      \\
\Checkmark & \Checkmark &                           & 91.1         & 90.0      \\
\Checkmark &                           & \Checkmark & 91.4            & 88.6         \\
                          & \Checkmark & \Checkmark & 91.6         & 86.3      \\
\Checkmark & \Checkmark & \Checkmark & \textbf{91.7} & \textbf{90.3} \\ \hline
\end{tabular}
\end{table}

Finally, in our last set of experiments, we compared the results of combining features at different aspects on finetuning and zero-shot classification tasks. Table 5 summarizes the results. Our findings showed that training strategies based on local and instance learning significantly improved zero-shot learning performance by establishing fine-grained associations between textual and visual elements. On the other hand, methods relying on modality training performed better in finetuning tasks but showed a decline in zero-shot performance. This is because these methods primarily focused on the representation capabilities of individual modalities, neglecting the intermodal correlations. In conclusion, training strategies designed based on different aspects offer distinct advantages. To achieve optimal performance, our approach combines the strengths of these methods, leveraging the benefits of each aspect to maximize overall model performance.

\section{Conclusion}

In this study, we proposed a novel multi-scale feature learning method for building medical vision-language foundation models. By effectively incorporating local, instance, modality, and global scale information, our method enhances the model's representation capabilities and performance across various clinical tasks. To evaluate the efficacy of our method, we conducted extensive experiments on six open-source datasets, covering diverse clinical tasks such as classification, segmentation, zero-shot classification, and phase grounding. The experimental results comprehensively validated the effectiveness of our method. The findings of this work hold great potential to contribute to the advancement of automated diagnostic systems and medical image analysis, offering new insights into achieving more accurate and comprehensive analysis of medical data in clinical practice.

\clearpage
\section{Compliance with ethical standards}
\label{sec:ethics}
This research study was conducted retrospectively using human subject data made available in open access. Ethical approval was not required as confirmed by the license attached with the open access data.

\section{Acknowledgments}
\label{sec:acknowledgments}
This research was partly supported by the National Natural Science Foundation of China (62222118, U22A2040), Shenzhen Science and Technology Program (RCYX20210706092-104034, JCYJ20220531100213029), Guangdong Provincial Key Laboratory of Artificial Intelligence in Medical Image Analysis and Application (2022B1212010011), the major key project of Peng Cheng Laboratory under grant PCL2023AS1-2, and Key Laboratory for Magnetic Resonance and Multimodality Imaging of Guangdong Province (2020B1212060051). 

\setstretch{0.5}
\bibliographystyle{IEEEbib}
\bibliography{strings,refs}

\begin{thebibliography}{10}

\bibitem{chang2023mining}
Qi~Chang et~al.,
\newblock ``Mining multi-center heterogeneous medical data with distributed synthetic learning,''
\newblock {\em Nature Communications}, vol. 14, no. 1, pp. 5510, 2023.

\bibitem{wang2021annotation}
Shanshan Wang et~al.,
\newblock ``Annotation-efficient deep learning for automatic medical image segmentation,''
\newblock vol. 12, no. 1, pp. 5915, 2021.

\bibitem{zhou2019d}
Yongjin Zhou et~al.,
\newblock ``{D-UN}et: a dimension-fusion u shape network for chronic stroke lesion segmentation,''
\newblock {\em IEEE/ACM transactions on computational biology and bioinformatics}, vol. 18, no. 3, pp. 940--950, 2019.

\bibitem{zhou2023transformer}
Hong-Yu Zhou et~al.,
\newblock ``A transformer-based representation-learning model with unified processing of multimodal input for clinical diagnostics,''
\newblock {\em Nature Biomedical Engineering}, pp. 1--13, 2023.

\bibitem{zhou2023unified}
Hong-Yu Zhou et~al.,
\newblock ``A {U}nified {V}isual {I}nformation {P}reservation {F}ramework for {S}elf-supervised {P}re-{T}raining in {M}edical {I}mage {A}nalysis,''
\newblock {\em IEEE Transactions on Pattern Analysis and Machine Intelligence}, vol. 45, no. 7, pp. 8020--8035, 2023.

\bibitem{krishnan2022self}
Rayan Krishnan, Pranav Rajpurkar, and Eric~J Topol,
\newblock ``Self-supervised learning in medicine and healthcare,''
\newblock {\em Nature Biomedical Engineering}, vol. 6, no. 12, pp. 1346--1352, 2022.

\bibitem{tang2022self}
Yucheng Tang et~al.,
\newblock ``Self-{S}upervised {P}re-{T}raining of {S}win {T}ransformers for 3d {M}edical {I}mage {A}nalysis,''
\newblock in {\em 2022 IEEE/CVF Conference on Computer Vision and Pattern Recognition (CVPR)}, 2022, pp. 20698--20708.

\bibitem{REFERS}
Hong-Yu Zhou et~al.,
\newblock ``Generalized radiograph representation learning via cross-supervision between images and free-text radiology reports,''
\newblock {\em Nature Machine Intelligence}, vol. 4, no. 1, pp. 32--40, 2022.

\bibitem{MRM}
Hong-Yu Zhou et~al.,
\newblock ``Advancing radiograph representation learning with masked record modeling,''
\newblock {\em The Eleventh International Conference on Learning Representations.}, 2022.

\bibitem{BIOVIL}
Benedikt Boecking et~al.,
\newblock ``Making the most of text semantics to improve biomedical vision-language processing,''
\newblock in {\em European conference on computer vision}. Springer, 2022, pp. 1--21.

\bibitem{GLORIA}
Shih-Cheng Huang et~al.,
\newblock ``{GL}o{RIA}: A {M}ultimodal {G}lobal-{L}ocal {R}epresentation {L}earning {F}ramework for {L}abel-efficient {M}edical {I}mage {R}ecognition,''
\newblock in {\em 2021 IEEE/CVF International Conference on Computer Vision (ICCV)}, 2021, pp. 3922--3931.

\bibitem{M-FLAG}
Che Liu et~al.,
\newblock ``{M-FLAG}: {M}edical vision-language pre-training with frozen language models and latent space geometry optimization,''
\newblock in {\em International Conference on Medical Image Computing and Computer-Assisted Intervention}. Springer, 2023, pp. 637--647.

\bibitem{MEDKLIP}
Chaoyi Wu et~al.,
\newblock ``Med{KLIP}: Medical {K}nowledge {E}nhanced {L}anguage-{I}mage {P}re-{T}raining,''
\newblock {\em Proceedings of the IEEE/CVF International Conference on Computer Vision}, 2023.

\bibitem{CONVIRT}
Yuhao Zhang et~al.,
\newblock ``Contrastive learning of medical visual representations from paired images and text,''
\newblock in {\em Machine Learning for Healthcare Conference}. PMLR, 2022, pp. 2--25.

\bibitem{MED-UNIC}
Zhongwei Wan et~al.,
\newblock ``Med-unic: {U}nifying cross-lingual medical vision-language pre-training by diminishing bias,''
\newblock {\em Advances in Neural Information Processing Systems}, vol. 36, 2024.

\bibitem{MaCo}
Weijian Huang et~al.,
\newblock ``Enhancing representation in radiography-reports foundation model: A granular alignment algorithm using masked contrastive learning,''
\newblock {\em arXiv preprint arXiv:2309.05904}, 2023.

\bibitem{CLIP}
Alec Radford et~al.,
\newblock ``Learning transferable visual models from natural language supervision,''
\newblock in {\em International conference on machine learning}. PMLR, 2021, pp. 8748--8763.

\bibitem{ALBEF}
Junnan Li et~al.,
\newblock ``Align before fuse: Vision and language representation learning with momentum distillation,''
\newblock {\em Advances in neural information processing systems}, vol. 34, pp. 9694--9705, 2021.

\bibitem{MGCA}
Fuying Wang et~al.,
\newblock ``Multi-granularity cross-modal alignment for generalized medical visual representation learning,''
\newblock {\em Advances in Neural Information Processing Systems}, vol. 35, pp. 33536--33549, 2022.

\bibitem{MAE}
Kaiming He et~al.,
\newblock ``Masked autoencoders are scalable vision learners,''
\newblock in {\em Proceedings of the IEEE/CVF conference on computer vision and pattern recognition}, 2022, pp. 16000--16009.

\bibitem{BERT}
Jacob Kenton et~al.,
\newblock ``{BERT}: Pre-training of {D}eep {B}idirectional {T}ransformers for {L}anguage {U}nderstanding,''
\newblock in {\em Proceedings of naacL-HLT}, 2019, vol.~1, p.~2.

\end{thebibliography}

\end{document}